# Investor reaction to financial disclosures across topics: An application of latent Dirichlet allocation

Feuerriegel, Stefan, ETH Zurich, Zurich, Switzerland, sfeuerriegel@ethz.ch

Pröllochs, Nicolas, University of Freiburg, Freiburg, Germany, nicolas.proellochs@is.uni-freiburg.de

**Abstract:**

This paper provides a holistic study of how stock prices vary in their response to financial disclosures across different topics. Thereby, we specifically shed light into the extensive amount of filings for which no a priori categorization of their content exists. For this purpose, we utilize an approach from data mining – namely, latent Dirichlet allocation – as a means of topic modeling. This technique facilitates our task of automatically categorizing, ex ante, the content of more than 70,000 regulatory 8-K filings from U.S. companies. We then evaluate the subsequent stock market reaction. Our empirical evidence suggests a considerable discrepancy among various types of news stories in terms of their relevance and impact on financial markets. For instance, we find a statistically significant abnormal return in response to earnings results and credit rating, but also for disclosures regarding business strategy, the health sector, as well as mergers and acquisitions. Our results yield findings that benefit managers, investors and policy-makers by indicating how regulatory filings should be structured and the topics most likely to precede changes in stock valuations.

# 1 Introduction

The efficient market hypothesis stipulates that investors consider all available information in their decision-making process and then adapt their trading accordingly (Fama et al., 1969; Fama, 1970). In this context, regulations usually require companies and managerial bodies to publish and distribute novel information via standardized channels in order to guarantee that each piece of information is equally available to all market participants. For instance, listed companies in the United States are legally obligated to release disclosures in the form of so-called 8-K filings via the Securities and Exchange Commission (SEC).

Financial markets incorporate information in both quantitative and qualitative forms such as the semantic content of financial disclosures (Loughran and McDonald, 2016). This written content represents a rich source of information and thus adds to the explanation of stock price dynamics as a response to financial events or accounting information. Examples of market-relevant information are addressed in studies that investigate how financial markets respond to the language in newspaper articles (Tetlock, 2007; Tetlock et al., 2008), earnings reports (Loughran and McDonald, 2011) or other types of regulatory disclosures (Hanley and Hoberg, 2012).

Disclosures, such as Form 8-K filings in the U.S., are mandated to adhere to a predefined structure. The underlying objective is to increase the ease with which the filings can be processed by investors. To facilitate this goal, the regulator defines specific sections in which information of a certain type has to be filed.[1] Examples belonging to specific firm events are section 2.01, reporting the completion of asset acquisitions, or section 2.02, containing the results of operations and financial conditions. For instance, Lerman and Livnat (2010) document market reactions (in abnormal stock returns, return volatility and trading volume) to specific sections in 8-K filings. Similarly, Hanley and Hoberg (2012) study how the risk-related sections in prospectuses from initial public offerings coincide with future litigation costs. However, in practice, the majority of filings cannot make use of the aforementioned structure, as their materials do not match the given subjects: instead, firms place their content in section 8 (which incorporates all other events that have not been covered by the predefined sections) or section 9 (for additional appendices). In fact, our analysis later reveals that 22.74 % of all disclosures contain materials classified under section 8 and 77.36 % contain optional appendices. Hence, rather than focusing on specific sections, it is necessary to perform a holistic study, including the content from the free-text fields.

---

1 See for example the current section rules for Form 8-K filings from the U.S. Securities and Exchange Commission detailed at https://www.sec.gov/fast-answers/answersform8khtm.html.

Yet little is known about which information – especially with regard to the unstructured materials in sections 8 and 9 – is relevant. Studying the relevance of individual sections within filings merely answers the question of what aspects of an event are informative to financial markets. However, such a study cannot adequately determine which events are relevant in the first place. In this sense, we hypothesize that not all news stories share the same degree of relevance, as some might convey anticipated information that has already been incorporated into the market or previously been released to the press, while others might not be decisive with regard to the future performance of firms.

This paper contributes to the existing body of research by performing an empirical, holistic study that dissects information reception with the help of data mining. The data for this study consists of regulated 8-K filings from companies listed on the New York Stock Exchange. We then categorize these disclosures according to their underlying content via so-called topic modeling. For this purpose, we propose the use of *latent Dirichlet allocation* (LDA) and compare the extracted topics from the 8-K filings to their average price reaction. This allows us to identify those topics that are of relevance for investors. Altogether, our findings indicate that news reception varies strongly across the different subjects of financial disclosures.

This piece of research reveals considerable implications for investors, management and communication departments. First of all, our work exposes which news items really matter to the audience and thus might be more relevant than others. We thus contribute to the perception and communication of information (McKinney and Yoos, 2010). Based on our findings, management and communication officers can now devote more time to these specific topics and thus enhance their communication with regard to relevant issues. On the other hand, practitioners can reduce effort spent on less relevant items and, for instance, find more standardized ways of writing and releasing these news stories. Above all, our work allows management – and especially investor relations officers – to more easily anticipate the likeliest direction of stock price changes subsequent to their news announcements. This knowledge thus helps to reduce uncertainty and gain a better understanding of the consequences of publicizing specific news stories. Finally, we make suggestions for policy-makers on how to align the current structure of regulatory filings with our observations of actual reporting needs.

The remainder of this paper is structured as follows. Section 2 provides an overview of related works that utilize data mining to gain insights into news reception across different content themes, while also deriving the relevance and novelty of our research question. Section 3 then presents our data mining methodology in order to empirically measure the reception of financial disclosures across different topics. Finally, the corresponding results are presented in Section 4, while Section 5 discusses implications for professionals, management and research in the fields of IS, data mining and decision sciences.

## 2  Background

This section summarizes the reception of financial materials by stock markets with a specific focus on previous efforts that also utilize data mining techniques. For reasons of conciseness, an extensive theoretical foundation in light of the efficient market hypothesis is given in the online appendix, which also includes a background pertaining to regulatory communication of corporates.

The content of regulatory announcements have been found as a driver for changes in future expectations and thus stock valuations (e. g. Loughran and McDonald, 2011, 2016). This relationship has further been studied with regard to different parts of 8-K filings (Lerman and Livnat, 2010) and by different topics which presents the focus of this paper. Here related studies typically investigate the effect of a single, specific disclosure topic on the stock market, such as share issues, changes in corporate governance, mergers and acquisitions (e. g. Chan, 2003; Tetlock, 2007; Vuolteenaho, 2002). However, the above studies only focus on one specific theme at a time and ignore the large body of disclosures that do not belong to any of the given topics, whereas we later present an approach by which perform a holistic analysis.

There exists another venue of research that extracts multiple topics from a collected body of publications and studies their influence on stock valuation. For instance, previous research classifies newspaper headlines into 20 predefined world-event categories in order to examine the effect of news on stock price movements (Niederhoffer, 1971). This study reveals a general tendency of stock prices to rise following positive newspaper headlines. However, it detects no statistically significant difference in price movements across different categories of newspaper headlines. A potential reason originates from the fact that markets incorporate new information very quickly and the novelty of information contained within frequently delayed newspapers is relatively low.

Closest to our research is the study by Neuhierl et al. (2013), which investigates the influence of new topics on abnormal returns. For this purpose, the authors manually divide a large corpus of corporate press releases into 10 main and 60 subgroups. They find that volatility tends to increase in the aftermath of a news release as a possible result of valuation uncertainty. In particular, the authors find significant reactions to news that is related to corporate strategies, customers and partners, products and services, management changes and legal documents.

A similar research paper also uses an *ex ante* list of 14 predefined categories like *acquisition*, *deal*, *legal* or *award* (Boudoukh et al., 2013). News items are then classified into these categories via rules from computational linguistics. Here the authors find that a few particular topics, such as *analyst recommendations* and *financials*, result in days of extreme stock price movements. Similarly, Feuerriegel et al. (2016) study topic-dependent reception of news for German ad hoc announcements, finding differences in the market response. In a different study, a Naïve Bayes algorithm is trained using a pre-labeled dataset that consists of articles from the *Wall Street Journal* (Antweiler and Frank, 2006). The labeled topics then serve as the basis from which to study the reaction of the stock market. Among others, this study observes the tendency of the market to overreact, as abnormal returns show opposite signs before and after publication. Moreover, the paper also reports a more prolonged impact of news stories during a recession as opposed to an expansion.

As a major drawback, the approaches previously listed in the literature section are typically driven by an *ex ante* list of given topics. In contrast, we extract topics *ex post* to match our corpus. We thus implement an automated data mining procedure that infers the topics from the language itself in a computerized fashion. For instance, this methodology has previously been utilized, for instance, to discover risk types in textual disclosures (Bao and Datta, 2014). By using the latent Dirichlet allocation from data mining to extract topics, we benefit from three advantages: first, we avoid a subjective bias from manual topic extraction and, second, we allow for greater flexibility for the topic selection method to match our corpus. As a matter of fact, we can even categorize the content in section 8 and 9 of 8-K filings into different different themes. In addition, the data mining approach enables us to process thousands of disclosures, thereby providing a comprehensive picture of news reception. Such an endeavor would be prohibitively difficult and costly with manual labeling.

It is worth noting that most of the textual information in SEC-regulated 8-K filings does not feature an explicit label denoting the topic of the content. This is particularly true of the extensive appendices, e. g., press releases, which are typically attached without specific information regarding their subject. As an example, 22.60 % of disclosures include section 8 ("other events") and 77.36 % have additional materials attached to them in the form of an appendix. To overcome this gap, we extract the corresponding topics using data mining in the following.

Accordingly, we later contribute to the literature by addressing our research questions: what topics are covered by 8-K filings? Which topics trigger an abnormal return of above/below zero? Which topics feature an above-average volatility? Based on our empirical findings, we can then report an array of implications for researchers, managers and policy-makers. In particular, we compare our identified topics to the prescribed structure of 8-K filings in order to derive suggestions on how to align both.

## 3 Methodology

This section details our methodology (see Figure 1) in order to investigate differences in the reception of financial filings. First of all, we apply several preprocessing steps to transform the written materials into a mathematical structure. This format then serves as input for the subsequent topic modeling by means of the latent Dirichlet allocation. Finally, we perform an event study in order to calculate the abnormal return corresponding to each publication.

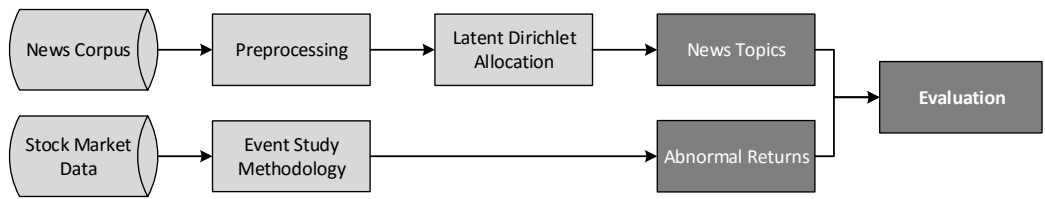

**Figure 1.** Methodology studying variations in information reception across different topics in financial disclosures.

### 3.1 Text Mining Procedure

Before we can carry out the topic modeling, we apply several preprocessing steps that are common in text mining (Manning and Schütze, 1999). These operations transform a running text into a matrix notation that allows for further calculations.

First of all, we remove terms (i. e. stop words) that frequently occur in the English language, such as *the*, *is* or *of*. These are unlikely to contribute to the meaning of the content and can thus be omitted. Here, we use a list of 174 stop words (Feinerer et al., 2008). We then map related words together by reducing inflected words to their stems. For this task, we utilize the so-called Porter stemming algorithm. The next step is to create a document-term matrix which stores the frequencies of each stem that occurs in the document collection. Since document-term matrices exhibit the full complexity of natural language, they can easily grow very large in size but also become very sparse, i. e. contain many zero entries. We thus drop uncommon words that appear in less than 5 % of all filings.

## 3.2 Topic Modeling via Latent Dirichlet Allocation

In this paper, we employ a state-of-the-art method from the text mining discipline for topic modeling, namely, the *latent Dirichlet allocation* (LDA). The LDA is a generative probabilistic model for extracting topics from a corpus (Blei, 2012). In its representation, it assumes that every document contains a mixture of hidden (i. e. latent) topics. The latter are defined as a probability distribution over the vocabulary. Based on a pre-defined number of topics, we can then infer a topic by choosing that which has the highest probability given a specific set of words. Accordingly, every document is assumed to have been generated by the following two-stage process (Blei et al., 2003; Blei, 2012):

1. For every document $d$ in corpus $D$, one draws a random distribution $\theta_d$ of topics, where entry $\theta_{d,k}$ gives the proportion of topic $k$ in $d$. The random variable $\theta_d$ follows a Dirichlet distribution with prior $\alpha$ given by

$$\theta_d \sim \text{Dir}(\alpha), \qquad \alpha = (\alpha_1, \alpha_2, \ldots, \alpha_K). \tag{1}$$

2. We now specify how words map onto topics. We thus select a topic $z_{d,n}$ from $\theta_d$ for every word $n$ in document $d$. Furthermore, one chooses a word $w_{d,n}$ over a fixed vocabulary conditioned on the chosen topic $z_{d,n}$. The corresponding distribution is given by $\beta_k \sim \text{Dir}(\eta)$ with prior $\eta$.

The joint likelihood is then

$$P(\theta,\beta,w,z) = \prod_{d=1}^{D} P(\theta_d \mid \alpha) \prod_{k=1}^{K} P(\beta_k \mid \eta) \prod_{n=1}^{N} P(z_{d,n} \mid \theta_d) P(w_{d,n} \mid z_{d,n}). \qquad (2)$$

Unfortunately, it is impossible to directly maximize the joint probability as one can merely observe documents and not topics. As remedy, the LDA sets out to find the highest posterior distribution. The posterior distribution

$$P(\theta,\beta,z \mid w,\alpha,\eta) = \frac{P(\theta,\beta,w,z \mid \alpha,\eta)}{P(w \mid \alpha,\eta)} \qquad (3)$$

is then obtained from dividing the joint likelihood by the marginal probability. However, the denominator is generally computationally intractable because of the dependence between $\beta$ and $\theta$. For this reason, one refers to approximate inference techniques like variational expectation-maximization (Steyvers and Griffiths, 2013).[2] In order to perform the LDA, we have to choose Dirichlet priors $\alpha$ and $\eta$ that control document-topic and topic-word distributions, respectively. We initialize all LDA parameters by following the default values used in the original paper by Blei et al. (2003).

---

2  Another common alternative is Gibbs sampling, which is a variation of Markov Chain Monte Carlo (MCMC).

In a next step, the latent Dirichlet allocation requires one to assign a unique identifier, i.e. topic name, to each of the extracted topics. To interpret a topic, one typically examines a ranked list of the 3 to 30 most probable terms in that topic. As a drawback, frequent and not decisive terms in the corpus commonly appear in such lists and, hence, render it difficult to differentiate the meanings of the topics. Consequently, recent research finds that ranking terms based on this probability hampers interpretation (e.g. Chang et al., 2009).

To mitigate this issue, we utilize the term-topic relationship scheme from Sievert and Shirley (2014) that facilitates topic interpretation by measuring the "relevance" of a term to a topic. As a main benefit, this method results in more coherent and interpretable topics (Sievert and Shirley, 2014). From a mathematical perspective, relevance is a weighted average of the logarithms of a term's probability within a topic $\phi_{k,n}$ and its "lift", where the "lift" is defined as the ratio of a term's probability within a topic $\phi_{k,n}$ to its marginal probability across the corpus $p_n$. The relevance score for word $n$ in topic $k$ is then calculated as

$$r(n,k\,|\,\lambda) = \lambda \log(\phi_{kn}) + (1-\lambda)\log\left(\frac{\phi_{kn}}{p_n}\right), \qquad (4)$$

where $\lambda$ is a weight given to topic *k* under word *n* relative to its lift. The relevance measure can be balanced with $0 \leq \lambda \leq 1$, by giving more weight to $\phi$ ($\lambda = 1$) or to the lift ($\lambda = 0$). In our study, we choose $\lambda = 0.6$, as suggested by Sievert and Shirley (2014) during an extensive analysis.

Finally, we need to determine how to assign a single topic to each announcement since, originally, the latent Dirichlet allocation determines a topic distribution $\theta_d$ for each disclosure *d*. Consistent with previous research, we assign each financial disclosure to the topic with the highest probability of occurrence.

### 3.3 Event Study Design

We now present our event study design (cf. MacKinlay, 1997; Srinivasan et al., 2017) in order to analyze the information value of a financial disclosure. This allows us to estimate the effect of a publication on the stock market without confounding influences. We then estimate a normal return in the absence of a disclosure and compare it to the observed return. The difference yields the abnormal return that can be attributed to the novel information from the financial disclosure entering the market. The appendix outlines the event study methodology in depth. We report all returns (in percent).

## 4   Empirical Findings

This section presents our empirical findings regarding an asymmetric information reception across different topics in financial disclosures. In this paper, we hypothesize that not all filings are equally relevant to the decision-making of investors. We thus aim at identifying those topics that trigger a significant adjustment of the stock price. For this purpose, we first present our underlying corpus of 8-K filings, followed by descriptive insights. We then automatically categorize the 8-K filings according to their subject using the latent Dirichlet allocation and study the average impact on stock prices by topic.

### 4.1   Corpus

We proceeded as follows in order to create our corpus of 8-K filings: we downloaded all 8-Ks, including headlines and amend documents, from the EDGAR website[3] from the years 2004 to 2013. This gives a total set of 901,133 filings, which then underwent several filtering steps. First, we select only filings from firms that are listed on the New York Stock Exchange (NYSE). Second, in order to gain information about the stock market reaction of investors, we remove filings for which we are not able match the SEC CIK numbers to NYSE stock symbols. Third, we exclude filings that contain fewer than 150 words (Loughran and McDonald, 2011). Consistent with the previous literature, we also remove penny stocks with a stock price below $5, as these frequently feature a higher volatility and thus may misrepresent the true magnitude of abnormal returns (Konchitchki and O'Leary, 2011). These filtering steps result in a final corpus of 73,986 filings.

---

3   U. S. Securities and Exchange Commission: http://www.sec.gov.

## 4.2 Descriptive Statistics

We now investigate the frequency and length of the filings in the following. Our corpus covers a total of 863 different companies, while the median number of filings per firm is 85.73 (with a standard deviation of 66.73). The total range goes from a minimum of 1 to a maximum of 501 for a single firm.

Table 1 shows descriptive statistics of the abnormal returns at the end of the trading day for the 8-K filings corpus. The mean change in abnormal returns for all 8-K filings in our sample is 0.0361 %. Out of all disclosures, a total of 36,587 resulted in a positive abnormal return, while 37,378 evoked a negative response. The abnormal stock market returns (in %) entail a standard deviation of 4.03 and median close to zero. Furthermore, Table 2 compares the characteristics of the corpus across each year. Evidently, the frequency and length of filings, as well as the number of covered firms, increases with each year.

|  | Mean | Median | Min. | Max. | Std. Dev. | Skewness | Kurtosis |
|---|---|---|---|---|---|---|---|
| Abnormal Return (in %) | 0.04 | −0.02 | −89.46 | 238.20 | 4.03 | 4.70 | 255.13 |
| Number of Filings (per Month) | 616.60 | 617.20 | 450.50 | 794.70 | 103.14 | 0.02 | −1.25 |
| Filing Length (Stemmed Words) | 3473.00 | 1001.00 | 151.00 | 648,700.00 | 10,018.06 | 15.01 | 523.89 |

**Table 1.** Descriptive statistics of the abnormal returns at the end of the first trading day (excluding penny stocks), as well as the length of the 8-K filings.

| Year | Average Abn. Ret. | Median Abn. Ret. | Std. Dev. Abn. Ret. | Total Filings | Mean Filing Length (Stemmed Words) | Covered Firms | Positive Filings | Negative Filings |
|---|---|---|---|---|---|---|---|---|
| 2004 | −0.030 | −0.063 | 3.21 | 5173 | 2611.25 | 479 | 2509 | 2664 |
| 2005 | 0.003 | −0.050 | 2.99 | 7066 | 2929.85 | 507 | 3427 | 3638 |
| 2006 | −0.033 | −0.053 | 3.07 | 7211 | 3059.41 | 517 | 3493 | 3718 |
| 2007 | 0.085 | −0.003 | 3.51 | 7197 | 3075.28 | 558 | 3588 | 3607 |
| 2008 | 0.064 | 0.041 | 6.21 | 7012 | 2861.81 | 546 | 3542 | 3467 |
| 2009 | 0.047 | −0.056 | 6.35 | 6700 | 3408.10 | 554 | 3293 | 3405 |
| 2010 | 0.014 | −0.011 | 3.28 | 7776 | 3645.02 | 601 | 3863 | 3913 |
| 2011 | 0.070 | 0.005 | 3.72 | 8110 | 3695.27 | 622 | 4064 | 4045 |
| 2012 | 0.040 | −0.026 | 3.51 | 8687 | 4356.49 | 649 | 4268 | 4409 |
| 2013 | 0.071 | 0.007 | 3.10 | 9054 | 4364.91 | 728 | 4540 | 4512 |

**Table 2.** Descriptive statistics of filings and abnormal returns (in %) across years (excluding penny stocks).

## 4.3 Topic Extraction

Regulated 8-K filings from stock-listed companies in the U. S. do not feature a code or label specifying the theme of their content. We overcome this gap by extracting the corresponding topics using the latent Dirichlet allocation. To perform this method, we have to choose ex ante the number of topics that we want to identify. This is different from other machine learning algorithms whereby one optimizes, for example, the number of clusters by cross-validation or heuristics. Concordant with related research, we run our analysis with 20 topics (Blei et al., 2003; Ramage et al., 2009; Niederhoffer, 1971). In addition, we also test a wide range, from merely 10 topics up to 40 topics, finding affirmative insights (see appendix for further details).

Finally, we assign an unique identifier, i. e. topic name, to each of the extracted topics. Specifically, we infer the individual topic names from the most relevant terms occurring in each given topic. For example, stemmed words, such as *director*, *appoint*, *vote*, *elect*, suggest a topic related to changes in management or corporate governance. On the other hand, words stems such as *gas*, *energy*, or *oil* represent disclosures that are related to the energy sector. A complete list of extracted topics and relevant terms is provided in the online appendix.

Table 3 provides descriptive statistics on the extracted topics, thereby indicating that the majority of documents are assigned to one of two topics, namely, *earnings* *results* and *public relations*, while the rest are distributed in a fairly even fashion across the remaining ones. Interestingly, the high share of, e. g., *earnings results* as a frequent topic in financial news is also consistent with findings from the existing literature (Carter and Soo, 1999). The table also shows that there are substantial differences regarding the filing lengths even across relatively similar news themes. For example, filings related to *income statements* exhibit an 80 % greater length in terms of included words as compared to *earnings results*. A thorough manual assessment of the corresponding filings reveals that *earnings results* typically contain a large proportion of highly compressed qualitative content, including particularly relevant business outlooks. In contrast, *income statements* consist of a larger amount of quantitative numbers, as well as less informative standard texts and regulatory notes.

|  | #Covered Firms | #Filings | Mean Filing Length (#Stemmed Words) |
|---|---|---|---|
| Energy Sector | 169 | 3779 | 1289.42 |
| Insurance Sector | 197 | 3761 | 2509.05 |
| Change of Trustee | 480 | 2075 | 14,644.95 |
| Real Estate | 149 | 2975 | 2897.59 |
| Corporate Structure | 540 | 1575 | 11,359.47 |
| Loan Payment | 567 | 1940 | 23,273.82 |
| Amendment of Shareholder Rights | 487 | 1738 | 3550.16 |
| Earnings Results | 672 | 14,246 | 1489.70 |
| Securities Sales | 496 | 2317 | 8039.96 |
| Stock Option Award | 670 | 5971 | 2748.03 |
| Credit Rating | 417 | 1246 | 4394.45 |
| Income Statements | 496 | 5592 | 2680.31 |
| Business Strategy | 581 | 6892 | 994.12 |
| Securities Lending | 326 | 1523 | 3368.34 |
| Management Change | 719 | 5105 | 1710.42 |
| Healthcare Sector | 83 | 507 | 6821.64 |
| Tax Report | 255 | 524 | 34,204.59 |
| Stock Dilution | 98 | 386 | 7706.84 |
| Mergers and Acquisitions | 295 | 612 | 13,167.53 |
| Public Relations | 699 | 11,222 | 409.99 |

**Table 3.** Descriptive statistics across all extracted topics.

## 4.4 Stock Market Response Across Disclosure Topics

We now investigate how the average stock market reaction to financial disclosures is related to the identified topics. For this purpose, we compare the distribution of abnormal returns for each topic. The farther away from a zero abnormal return, the more strongly a given topic is linked to an actual investment decision by investors. This can also be statistically tested by, e. g., a *t*-test.

According to Table 4, we observe the following overall pattern. First, the median abnormal returns are closely located around zero for most topics. However, several topics – such as *business strategy* – exhibit a positive median abnormal return, whereas *mergers and acquisitions* and *management change*, among others, show a negative impact on the market.

Table 4 also reflects the outcome of hypothesis testing. Here we bootstrap confidence intervals with $N = 200$ replacements due to the small number of observations. According to our results, a total number of five topics yield a *p*-value below the threshold of the 5 % statistical significance level. Specifically, we find a statistically significant effect in a negative direction for *earnings results* and *mergers and acquisitions*. These topics are linked to negative median abnormal returns of −0.01 % and −0.03 %, respectively. On the other hand, *business strategy*, *healthcare sector*, and *credit rating* are linked to positive market responses.

| No. | Topic Name | Abnormal Return | | | |
|---|---|---|---|---|---|
| | | Median | Abs. Median | p-Val. | Std. Dev. |
| 1 | Energy Sector | −0.01 | 1.07 | 0.71 | 2.86 |
| 2 | Insurance Sector | −0.03 | 1.11 | 0.21 | 4.10 |
| 3 | Change of Trustee | 0.01 | 0.87 | 0.36 | 2.62 |
| 4 | Real Estate | −0.07 | 1.04 | 0.84 | 2.45 |
| 5 | Corporate Structure | −0.08 | 1.21 | 0.10 | 6.71 |
| 6 | Loan Payment | −0.04 | 1.01 | 0.15 | 6.09 |
| 7 | Amendment of Shareholder Rights | −0.03 | 1.16 | 0.66 | 3.61 |
| **8** | **Earnings Results** | **−0.01** | **2.15** | **0.02** | **5.40** |
| 9 | Securities Sales | −0.02 | 0.97 | 0.90 | 2.63 |
| 10 | Stock Option Award | −0.04 | 1.02 | 0.30 | 2.66 |
| **11** | **Credit Rating** | **0.00** | **1.10** | **0.04** | **4.25** |
| 12 | Income Statements | 0.07 | 1.69 | 0.12 | 4.81 |
| **13** | **Business Strategy** | **0.04** | **1.20** | **0.01** | **3.40** |
| 14 | Securities Lending | −0.04 | 0.85 | 0.52 | 4.45 |
| 15 | Management Change | −0.08 | 0.94 | 0.90 | 2.57 |
| **16** | **Healthcare Sector** | **0.03** | **1.20** | **0.02** | **3.83** |
| 17 | Tax Report | −0.07 | 1.07 | 0.97 | 5.11 |
| 18 | Stock Dilution | 0.15 | 1.46 | 0.22 | 4.08 |
| **19** | **Mergers and Acquisitions** | **−0.03** | **1.28** | **0.05** | **3.12** |
| 20 | Public Relations | −0.04 | 1.08 | 0.87 | 3.10 |

**Table 4.** Summary statistics for abnormal return (in %) across all extracted topics. Bold font highlights topics that are statistically significant at the 5 % significance level. Due to the small number of observations, we perform bootstrapping with $N = 200$ replacements.

## 4.5 Asymmetric Reception

In order to investigate potential asymmetries regarding the reception of different news topics, we now focus on the bandwidth of abnormal returns. As some topics might lead to higher stock price movements on average, a higher volatility implicitly suggests a larger variety in information content for investors.

We start by analyzing how the absolute abnormal returns differ across the identified news topics. As shown in the second column of Table 4, the median absolute abnormal returns show a large spread, ranging from 0.85 % for *securities lending* to 2.15 % for *earnings results*. In order to test whether these differences between medians are statistically significant, we utilize a Kruskal-Wallis test. According to our results, the null hypothesis that the median absolute abnormal return is equal across all topics is strongly rejected at the 1 % significance level. Hence, the information content and attention that individual filings receive from investors cannot be assumed to be fixed, but rather depends on the news topic.

Next, we verify this finding using the standard deviation of the abnormal returns (in percent) as an alternative measure for the stock market volatility. Also here, we observe a large spread regarding the volatility of the abnormal stock market returns. The standard deviation for the individual topics ranges from 2.45 to 6.71. Interestingly, we find a smaller bandwidth for, e. g., *real estate*, which accounts for a standard deviation of 2.45, while, e. g., *corporate structure* (6.71) and *loan* *payment* (6.09) feature a higher variance and, therefore, a higher information level. Such a homogeneity of variances can also be tested statistically. For this purpose, we draw upon the Bartlett's test in order to investigate the null hypothesis that all topics are drawn from populations with equal variances. The null hypothesis is strongly rejected at the 1 % significance level. Thus, once again, we see that the information level of 8-K filings strongly differs across the individual topics.

### 4.6 Comparison to Previous Research

We now compare our findings to previous research. We thus discuss the topics with a statistically significant, non-zero influence on the stock market in the following.

- *Earnings results.* This topic yields the highest volatility among all the significant topics. Furthermore, we also find a negative median abnormal return for disclosures, which largely corresponds to previous research by Lucas and McDonald (1990). In this same vein, Loughran and McDonald (2011) studied regulatory earnings reports whose tone is linked to the subsequent stock return. Even hidden or difficult-to-attain materials, such as footnotes, in earnings reports are eventually reflected in market prices (Bloomfield, 2002). Li (2008) observed a clear link between the readability of annual reports and firm performance, further bolstering the notion that investors pay close attention to this subject. Altogether, this data suggests a decisive role of this topic in relation to the performance of stocks.

- *Credit rating.* Credit-related materials typically convey highly relevant information to the capital market regarding the value of the borrowing firm. Such materials have thus been found to explain movements in stock prices (Lummer and McConnell, 1989).

- *Business strategy.* The significant effect of *business strategy* is often impelled by positively connoted long-term projects (Woolridge and Snow, 1990); for instance, with regard to international operations (Elango et al., 2013). This type of information might be conveyed by the optional appendices containing press-related materials. Similarly, newspapers tend to discuss not individual financial figures from firms, but rather the overall outlook or strategy. Consistent with our results, newspaper articles have also been found to relate to stock price changes (Tetlock, 2007; Tetlock et al., 2008).

- *Healthcare sector.* Our results show that disclosures in the *healthcare sector* have a significantly positive effect on the stock market. Similar evidence has also recently been reported by Feuerriegel et al. (2016). Interestingly, this also coincides with the fact that the healthcare industry has performed above average over the last decade (Arouri and Nguyen, 2010).

- *Mergers and acquisitions.* In contrast, disclosures related to *mergers and acquisitions* are linked to a negative price change on average. According to Cartwright and Schoenberg (2006), bidding firms frequently experience share price under-performance during the acquisition of another company. This is in contrast to the findings of earlier works on mergers and acquisitions, which predominantly focused on the announcement day. For instance, the work of Dodd (1980) identifies a positive market reaction regarding the price of a target firm subsequent to the announcement of an intended merger. Analogously, target firms achieve significantly positive abnormal returns during these announcement day (Asquith et al., 1983). In contrast, our study monitors all events related to such a process and incorporates them into our dataset. Hence, our filings also reflect the various hurdles associated with a merger or acquisition, as well as a potential failure of the process.

Our above discussion emphasizes the need for accurately labeling events in regulatory filings, especially with regard to the different phases of mergers and acquisitions. However, the current structure of 8-K filings falls short of this goal and we thus put forth suggestions in the next section.

## 5 Implications

In the following, we discuss the implications of our research as it allows for a better comprehension of decision-making in a financial context. Accordingly, our work not only contributes to research in the field of decision sciences, but is also highly relevant for executives and professionals when releasing information on firm performance to the public.

## 5.1 Implications for Research

The above results contribute to an understanding of how humans process information encoded in natural language. Our findings provide evidence that the reception and responses of investors varies greatly according to the story in question. As such, investors seem to selectively focus their attention only on certain themes.

Moreover, we also provide an intriguing approach for researchers in the fields of data mining and decision sciences to replicate our methodology for better understanding decision-making. By utilizing methods for topic modeling, researchers can process large volumes of textual materials without prior knowledge. When linking these insights to exogenous variables, topics are ranked accordingly and one finds topics with the highest information value. This can be applied to numerous domains beyond finance, such as social media, blog posts or user-generated content in recommender systems – thereby broadening our knowledge of how information and especially natural language facilitates decision-making; see e. g. Sul et al. (2017); Lu et al. (2012).

A number of works in the field (e.,g. Henry, 2008; Loughran and McDonald, 2011; Tetlock, 2007; Tetlock et al., 2008) treat all financial disclosures alike, without considering a potential heterogeneity in the reception across topics. Our work, however, is positioned around recent analyses in finance-related research according to which the response to financial materials cannot be assumed to be independent of decision-making strategies (Schiffels et al., ming; Taylor, 1975). Instead, researchers must accurately identify and account for such confounding factors in order to achieve rigor.

## 5.2 Implications for Policy-Makers

Our findings open an avenue for policy-makers to create value from data mining techniques. Currently, regulatory filings in the U.S. and other countries are distributed via standardized channels but not in a standardized format. Hence, filtering by topic is hardly possible given the status quo. To alleviate this, one could require firms to assign topic codes to each filing in order to facilitate filtering. Moreover, additional prerequisites regarding the format could improve machine readability. In fact, our findings have already found their way into recent projects of policy-makers, such as the Financial Reporting Council. The Financial Reporting Council is the U.K. regulator for corporate reporting, whom we advised based on this research to strive for improved financial reporting by the above means.

Since natural language still remains a challenge for computers to understand correctly, also adding semantic annotations (or linked data) could greatly facilitate this task (e.g. Storey et al., 2008). In a next step, this form of structured information could also ease the process of information filtering for all investors (including automated traders). Simultaneously, semantic structures show a potential path to-wards decreasing uncertainty regarding meaning and the subjective interpretation of financial materials and indirectly leveraging transparency.

Our results also yield insights into how the structure of regulatory filings might be improved in the future. Our analysis reveals only a partial match between the topics identified in Table 4 and the actual sections mandated by the U. S. Securities and Exchange Commission. We observe that specific sectors – namely energy, insurance and healthcare – make over-proportional use of free-text fields. Two options might be suitable to better reflect the publication needs of these firms, as well as the processing capabilities of potential investors. One the one hand, it might be possible to augment regulatory filings with a sector code in order to signal the market area in which the firm operates (or which the disclosure relates to). On the other hand, this might serve as the starting point for a rigorous, in-depth content analysis whereby it is determined that the current section names match with the disclosed information.

We also see an overlap between sections and identified topics for a variety of issues. Examples include corporate structure (item 6.02 refers to a change of trustee), stock dilution (item 3.01 addresses delistings) and amendments (item 5.03). However, other topics entail a considerably more granular sectional structure than suggested by our analysis: management change (items 5.01 to 5.08 dissect individual actions such as changes in control of registrant to shareholder director nominations) and financial information (item 2.02 deals with operation results, item 2.03 with other obligations, etc.). This observation is supported by earlier literature on management changes

(e. g. Bonnier and Bruner, 1989; Warner et al., 1988). Similarly, our discussion in Section 4.6 highlights the importance of financial results for the performance of firms. The more granular structure also reflects the human categorization by Neuhierl et al. (2013).

Yet the following items might be prevalent enough to justify sections of their own: real estate (largely conveying expressions related to leases, estates or building size belonging to firms), tax report (e. g. section 2 could be extended by a corresponding item) or mergers and acquisitions. With regard to the latter, Neuhierl et al. (2013) specifically distinguish intent and target, as well as subforms. In addition, section 9 is frequently utilized for both financial statements and exhibits containing additional materials (i. e. largely press releases). Since one of our topics specifically refers to press releases, policy-makers could consider splitting this section into two separate parts, such that investors can easily distinguish them. In contrast to Neuhierl et al. (2013), our LDA does not point towards categories covering meetings/events, products/services and legal.

### 5.3 Implications for Management and Professionals

With regard to managerial implications, we draw our attention to how information management influences firm performance (Mithas et al., 2011). Our findings provide decision support for managers by addressing the question of which subjects actually matter to investors. This helps managers and communication professionals to prioritize in which press releases to invest the most effort. Simultaneously, they can use the above framework based on data mining techniques to monitor and understand the performance of other firms relative to their own as an approach to competitive intelligence (Pröllochs and Feuerriegel, 2018; Zheng et al., 2012).

On the other hand, our approach outlines potential gains for professionals who need to process large quantities of disclosures in order to identify relevant information, such as investors or media bodies. By utilizing the LDA for topic identification, one can easily automate filtering for information. This is especially helpful when topic codes are not available, as in the current setting. Overall, our work contributes to the existing body of research regarding how the narrative content of disclosures can provide decision support for investors (Feuerriegel and Prendinger, 2016; Kraus and Feuerriegel, 2017; Pröllochs et al., 2016; Schumaker and Chen, 2009).

## 5.4 Opportunities for Future Research

Our work opens an avenue for future research related to decision sciences. While we have mainly focused on information reception, it would be an interesting extension to study information requirements and information needs on an individual basis. The former represents information that is objectively necessary for a trading decision, while the latter denotes what is subjectively considered to be relevant (cf. Zimbra et al., 2015). With further advances in text mining, one might be able to analyze information demand (i.e. what is sought by investors in decision-making) and extract the relevant parts of disclosures. In addition, it is also an intriguing task to study the reasons that underlie the different characteristics of decision-making across individual topics.

# 6 Concluding Remarks

Investors modify their trading in the wake of financial disclosures, yet not all filings appear equally relevant. It is thus the objective of this paper to contribute a holistic study of 8-K filings across different subjects. For this purpose, we utilize the latent Dirichlet allocation, as it allows us to automatically analyze thousands of 8-K filings from U.S. companies with respect to their content. This especially allows us to categorize the three-quarters of filings for which the content does not match the given list of topic codes, i.e. entailing textual materials in rubrics labeled as "other events" or appendices.

Our analysis yields a list of 20 topics reflecting the themes with the greatest need for reporting by businesses. Furthermore, our results provide evidence that stock market participants process information asymmetrically: a set of 5 topics triggers abnormal returns that are different from zero at a statistically significant level. Based on these findings, executives and communication professionals can immediately leverage our findings and focus their efforts on the stories that are relevant to the audience. By correctly interpreting acquired information, investors can improve their decision-making when buying and selling in order to increase the performance of their portfolio (Oztekin et al., 2016). Policy-makers can bring the reporting structure in line with the actual needs of firms.

## References

Antweiler, W., and Frank, M. Z. 2006. "Do US Stock Markets Typically Overreact to Corporate News Stories?" *SSRN Electronic Journal* .

Arouri, M. E. H., and Nguyen, D. K. 2010. "Oil Prices, Stock Markets and Portfolio Investment: Evidence from Sector Analysis in Europe Over the Last Decade," *Energy Policy* (38:8), pp. 4528–4539.

Asquith, P., Bruner, R. F., and Mullins, D. W. 1983. "The Gains to Bidding Firms From Merger," *Journal of Financial Economics* (11:1-4), pp. 121–139.

Bao, Y., and Datta, A. 2014. "Simultaneously Discovering and Quantifying Risk Types from Textual Risk Disclosures," *Management Science* (60:6), pp. 1371–1391.

Blei, D. M. 2012. "Probabilistic Topic Models," *Communications of the ACM* (55:4), pp. 77–84.

Blei, D. M., Ng, A. Y., and Jordan, M. I. 2003. "Latent Dirichlet Allocation," *The Journal of Machine Learning Research* (3), pp. 993–1022.

Bloomfield, R. J. 2002. "The 'Incomplete Revelation Hypothesis' and Financial Reporting," *Accounting Horizons* (16:3), pp. 233–243.

Bonnier, K.-A., and Bruner, R. F. 1989. "An Analysis of Stock Price Reaction to Management Change in Distressed Firms," *Journal of Accounting and Economics* (11:1), pp. 95–106.

Boudoukh, J., Feldman, R., Kogan, S., and Richardson, M. 2013. "Which News Moves Stock Prices? A Textual Analysis," .

Carter, M. E., and Soo, B. S. 1999. "The Relevance of Form 8-K Reports," *Journal of Accounting Research* (37:1), p. 119.

Cartwright, S., and Schoenberg, R. 2006. "Thirty Years of Mergers and Acquisitions Research: Recent Advances and Future Opportunities," *British Journal of Management* (17:S1), pp. S1–S5.

Chan, W. S. 2003. "Stock Price Reaction to News and No-News: Drift and Reversal After Headlines," *Journal of Financial Economics* (70:2), pp. 223–260.

Chang, J., Boyd-Graber, J., Wang, C., Gerrish, S., and Blei, D. M. 2009. "Reading Tea Leaves: How Humans Interpret Topic Models," in *Advances in Neural Information Processing Systems (NIPS)*, pp. 288–296.

Dodd, P. 1980. "Merger Proposals, Management Discretion and Stockholder Wealth," *Journal of Financial Economics* (8:2), pp. 105–137.

Elango, B., Talluri, S. S., and Hult, G. T. M. 2013. "Understanding Drivers of Risk-Adjusted Performance for Service Firms with International Operations," *Decision Sciences* (44:4), pp. 755–783.

Fama, E. F. 1970. "Efficient Capital Markets: A Review of Theory and Empirical Work," *Journal of Finance* (25:2), pp. 383–417.

Fama, E. F., Fisher, L., Jensen, M. C., and Roll, R. 1969. "The Adjustment of Stock Prices to New Information," *International Economic Review* (10:1), p. 1.

Feinerer, I., Hornik, K., and Meyer, D. 2008. "Text Mining Infrastructure in R," *Journal of Statistical Software* (25:5), pp. 1–54.

Feuerriegel, S., and Prendinger, H. 2016. "News-based trading strategies," *Decision Support Systems* (90), pp. 65–74.

Feuerriegel, S., Ratku, A., and Neumann, D. 2016. "Analysis of How Underlying Topics in Financial News Affect Stock Prices Using Latent Dirichlet Allocation," in *49th Hawaii International Conference on System Sciences (HICSS)*, IEEE Computer Society, pp. 1072–1081.

Hanley, K. W., and Hoberg, G. 2012. "Litigation Risk, Strategic Disclosure and the Underpricing of Initial Public Offerings," *Journal of Financial Economics* (103:2), pp. 235–254.

Henry, E. 2008. "Are Investors Influenced By How Earnings Press Releases Are Written?" *Journal of Business Communication* (45:4), pp. 363–407.

Konchitchki, Y., and O'Leary, D. E. 2011. "Event Study Methodologies in Information Systems Research," *International Journal of Accounting Information Systems* (12:2), pp. 99–115.

Kraus, M., and Feuerriegel, S. 2017. "Decision support from financial disclosures with deep neural networks and transfer learning," *Decision Support Systems* (104), pp. 38–48.

Lerman, A., and Livnat, J. 2010. "The New Form 8-K Disclosures," *Review of Accounting Studies* (15:4), pp. 752–778.

Li, F. 2008. "Annual Report Readability, Current Earnings, and Earnings Persistence," *Journal of Accounting and Economics* (45:2), pp. 221–247.

Loughran, T., and McDonald, B. 2011. "When Is a Liability Not a Liability? Textual Analysis, Dictionaries, and 10-Ks," *Journal of Finance* (66:1), pp. 35–65.

Loughran, T. I., and McDonald, B. 2016. "Textual Analysis in Accounting and Finance: A Survey," *Journal of Accounting Research* (54:4), pp. 1187–1230.

Lu, H.-M., Tsai, F.-T., Chen, H., Hung, M.-W., and Li, S.-H. 2012. "Credit Rating Change Modeling Using News and Financial Ratios," *ACM Transactions on Management Information Systems* (3:3), pp. 1–30.

Lucas, D. J., and McDonald, R. L. 1990. "Equity Issues and Stock Price Dynamics," *Journal of Finance* (45:4), pp. 1019–1043.

Lummer, S. L., and McConnell, J. J. 1989. "Further Evidence on the Bank Lending Process and the Capital-Market Response to Bank Loan Agreements," *Journal of Financial Economics* (25:1), pp. 99–122.

MacKinlay, A. C. 1997. "Event Studies in Economics and Finance," *Journal of Economic Literature* (35:1), pp. 13–39.

Manning, C. D., and Schütze, H. 1999. *Foundations of Statistical Natural Language Processing*, Cambridge, MA: MIT Press.

McKinney, E. H., Jr., and Yoos, C. J., II 2010. "Information About Information: A Taxonomy of Views," *MIS Quarterly* (34:2), pp. 329–344.

Mithas, S., Ramasubbu, N., and Sambamurthy, V. 2011. "How Information Management Capability Influences Firm Performance," *MIS Quarterly* (35:1), pp. 237–256.

Neuhierl, A., Scherbina, A., and Schlusche, B. 2013. "Market Reaction to Corporate Press Releases," *Journal of Financial and Quantitative Analysis* (48:04), pp. 1207–1240.

Niederhoffer, V. 1971. "The Analysis of World Events and Stock Prices," *Journal of Business* (44:2), pp. 193–219.

Oztekin, A., Kizilaslan, R., Freund, S., and Iseri, A. 2016. "A Data Analytic Approach to Forecasting Daily Stock Returns in an Emerging Market," *European Journal of Operational Research* (253:3), pp. 697–710.

Pröllochs, N., and Feuerriegel, S. 2018. "Business analytics for strategic management: Identifying and assessing corporate challenges via topic modeling," *Information & Management* (forthcoming).

Pröllochs, N., Feuerriegel, S., and Neumann, D. 2016. "Negation scope detection in sentiment analysis: Decision support for news-driven trading," *Decision Support Systems* (88), pp. 67–75.

Ramage, D., Hall, D., Nallapati, R., and Manning, C. D. 2009. "Labeled LDA: A Supervised Topic Model for Credit Attribution in Multi-labeled Corpora," in *Proceedings of the 2009 Conference on Empirical Methods in Natural Language Processing*, EMNLP '09, Stroudsburg, PA, USA: Association for Computational Linguistics, pp. 248–256.

Schiffels, S., Fliedner, T., and Kolisch, R. forthcoming. "Human Behavior in Project Portfolio Selection: Insights from an Experimental Study," *Decision Sciences* .

Schumaker, R. P., and Chen, H. 2009. "A Quantitative Stock Prediction System Based on Financial News," *Information Processing & Management* (45:5), pp. 571–583.

Sievert, C., and Shirley, K. E. 2014. "LDAvis: A Method for Visualizing and Interpreting Topics," in *Proceedings of the Workshop on Interactive Language Learning, Visualization, and Interfaces*, pp. 63–70.

Srinivasan, A., Guo, H., and Devaraj, S. 2017. "Who Cares About Your Big Day? Impact of Life Events on Dynamics of Social Networks," *Decision Sciences* (48:6), pp. 1062–1097.

Steyvers, and Griffiths 2013. "Probabilistic Topic Models," in *Handbook of Latent Semantic Analysis*, T. K. Landauer, (ed.), Mahwah, N.J.: Psychology Press, pp. 424–440.

Storey, V. C., Burton-Jones, A., Sugumaran, V., and Purao, S. 2008. "CONQUER: A Methodology for Context-Aware Query Processing on the World Wide Web," *Information Systems Research* (19:1), pp. 3–25.

Sul, H. K., Dennis, A. R., and Yuan, L. I. 2017. "Trading on Twitter: Using Social Media Sentiment to Predict Stock Returns," *Decision Sciences* (48:3), pp. 454–488.

Taylor, R. N. 1975. "Psychological Determinants of Bounded Rationality: Implications for decision-making strategies," *Decision Sciences* (6:3), pp. 409–429.

Tetlock, P. C. 2007. "Giving Content to Investor Sentiment: The Role of Media in the Stock Market," *Journal of Finance* (62:3), pp. 1139–1168.

Tetlock, P. C., Saar-Tsechansky, M., and Macskassy, S. 2008. "More Than Words: Quantifying Language to Measure Firms' Fundamentals," *Journal of Finance* (63:3), pp. 1437–1467.

Vuolteenaho, T. 2002. "What Drives Firm-Level Stock Returns?" *Journal of Finance* (57:1), pp. 233–264.

Warner, J. B., Watts, R. L., and Wruck, K. H. 1988. "Stock Prices and Top Management Changes," *Journal of Financial Economics* (20), pp. 461–492.

Woolridge, J. R., and Snow, C. C. 1990. "Stock Market Reaction to Strategic Investment Decisions," *Strategic Management Journal* (11:5), pp. 353–363.

Zheng, Z., Fader, P., and Padmanabhan, B. 2012. "From Business Intelligence to Competitive Intelligence: Inferring Competitive Measures Using Augmented Site-Centric Data," *Information Systems Research* (23:3), pp. 698–720.

Zimbra, D., Chen, H., and Lusch, R. F. 2015. "Stakeholder Analyses of Firm-Related Web Forums," *ACM Transactions on Management Information Systems* (6:1), pp. 1–38.